\documentclass{llncs}
\usepackage{hyperref}
\usepackage[utf8]{inputenc}
\usepackage{url}
\usepackage{tikz}
\usetikzlibrary{calc,positioning}
\usetikzlibrary{decorations.text}
\usetikzlibrary{arrows.meta}
\usetikzlibrary{decorations,backgrounds, shapes, shadows}
\usepackage{tikzpeople}
\usepackage{pgfplots}
\usepackage{amsmath,amsfonts}
\usepackage[stable]{footmisc}
\usepackage{authblk}
\usepackage{xcolor}
\usepackage{graphicx}
\graphicspath{{./Figures/}}
\usepackage[shortlabels]{enumitem}
\usepackage{multicol}
\usepackage{multirow}
\usepackage{comment}
\usepackage[numbers]{natbib}
\usepackage{todonotes}

\def\paragraph#1{\subsubsection{#1}}

\title{On some foundational aspects of\\ human centered Artificial Intelligence}
\author{Luciano Serafini\inst{1} \and
Raul Barbosa \inst{2} \and
Jasmin Grosinger\inst{3} \and 
Luca Iocchi \inst{4} \and
Christian Napoli\inst{4} \and
Salvatore Rinzivillo\inst{5} \and 
Jacques Robin\inst{6} \and 
Alessandro Saffiotti\inst{3} \and 
Teresa Scantamburlo \inst{7} \and
Peter Sch\"uller\inst{8} \and
Paolo Traverso\inst{1} \and 
Javier V\'azquez-Salceda\inst{9}}

\institute{Fondazione Bruno Kessler, Trento, Italy \and 
University of Coimbra, Portugal \and 
\"Orebro University, Sweden \and
Universit\'a di Roma ``La Sapienza'', Italy \and 
Istituto di Scienza e Tecnologie dell’Informazione, National Research
Council, Italy \and 
Universit\'e Paris 1 Panthon-Sorbonne, France \and
Ca' Foscari University, Italy \and
Technische Universit\"at Wien, Austria \and
Universitat Politecnica de Catalunya, Spain
}

\begin{document}
\maketitle


\begin{abstract}
  The burgeoning of AI has prompted recommendations that AI techniques
  should be ``human-centered''.  However, there is no clear definition
  of what is meant by Human Centered Artificial Intelligence, or for
  short, \emph{HCAI}.  This paper aims to improve this situation by
  addressing some foundational aspects of
  HCAI.  To do so, we introduce the term \emph{HCAI agent} to refer to
  any physical or software computational agent equipped with AI
  components and that interacts and/or collaborates with humans.  This
  article identifies five main conceptual components that participate in
  an HCAI agent: \emph{Observations}, \emph{Requirements},
  \emph{Actions}, \emph{Explanations} and \emph{Models}.
  We see the notion of HCAI agent, together with its components and
  functions, as a way to bridge the technical and non-technical
  discussions on human-centered AI.
  In this paper, we focus our analysis on scenarios consisting of a
  single agent operating in dynamic environments in presence of humans.
  %
%
  \end{abstract}



\section{Introduction}

Recently artificial intelligence (AI) systems become popular and are
playing an increasingly important and pervasive role in the life of
individuals and society.  The fact that AI systems can significatively
affect human lives, and can play an active role in society, is fostering
discussions on how these new systems should behave in their relationship
with humans.  Human Centered AI (HCAI) concerns the study of how present
and future AI systems will interact with human lives in a mixed society
composed of artificial and human agents, and how to keep the human into
the focus in this mixed society.
%
Due the interdisciplinarity of the topic, the discussion involves both
technical and non technical people which see AI agents from different
perspectives.  It is often difficult to integrate these perspectives in
a coherent vision in which all points of view can contribute and
complement each other.  Of particular interests are the recent attempts
to provide requirements and ethical guidelines that regulate HCAI
systems, like the ones published by the European Commission in
2019~\cite{eu-ethical-guidelines}.  While such guidelines will certainly
have an important impact on the future of AI, they typically describe
HCAI systems at a very abstract, qualitative level.  Understanding the
concrete impact of such abstract requirements on the technical
development of HCAI systems is neither trivial nor unique.  A case in
point is the recent European legislative proposal for an ``AI
Act''~\cite{AIAct.ec2021}.  The necessity to define a precise grounding
of high level qualitative statements into technical requirements for
HCAI systems was explicitly noted in the above EC guidelines:

\begin{quote}
Requirements for Trustworthy AI should be ''translated'' into
procedures and/or constraints on procedures, which should be anchored
in the AI system’s architecture. \cite[page 21]{eu-ethical-guidelines}.
\end{quote}


The main objective of this document is to provide a reference
description of an HCAI agent, together with its main components and
functions, that helps to bridge the gap between the abstract
specifications of HCAI systems provided by non-technical entities and
the real implementation of these systems.  Such a description is
intended to contribute to the above grounding.


In other words, the main objective of this paper is an attempt to fill
the gap between ``non technical description of HCAI agent'' and more
``technical structure'' that is understandable without a deep knowledge
of the main technics of AI, such as Machine Learning, Automated
Reasoning, Reinforcement Learning, Probabilistic inference and
optimisation.  Despite this, we will try to be precise enough to allow
the mapping of high level concepts that one can find for instance in the
EC ethical guidelines into a more technical and technically operative
description.  Although the account of HCAI agents that we offer in this
paper is certainly preliminary, we believe that it constitutes an
important first step towards filling a crucial cultural gap.

The rest of this paper is organized as follows.  In the next section, we
provide a definition of HCAI agents in terms of their neessary features.
In Section~\ref{schema.sec} we offer a schematic represesentation of a
HCAI agent, and in Section~\ref{keys.sec} we discuss the key components
of this representation.  In Section~\ref{models.sec} we look in more
detail at one of these components, AI models.
Finally, we discuss related work in Section~\ref{related.sec}, and draw
some concluding remarks in Section~\ref{conclusions.sec}.


\section{Definition of HCAI agent}
\label{hcai.sec}

In the following, we provide a connotative definition of an HCAI agent by
listing a set of features that an AI agent should meet in order to be
considered Human Centered.

\paragraph{Agent-environment pair.}

\def\agent{\mathbf{Ag}}
\def\agentState{State(\agent)}
\def\environment{\mathbf{Env}}
\def\environmentState{State(\environment)}

We consider a reference framework composed of an autonomus agent that
operates in an environment that includes the presence of humans.  We
call this an \emph{agent-environment pair}.  HCAI concentrates on
agent-environment pairs in which humans directly interact with the
artificial agent in a \emph{collaborative attitude}.  We therefore
exclude from our analysis situations in which the agent competes with
humans, such as the ones recently studied in \cite{fraune2019human}.  In
an agent-environment pair, there is a clear separation between what is
internal to the agent and what is external to the agent, i.e., the
environment.  At every time point in time, both the agent and the
enviroment are in a state, that we refer as the state of the agent and
the state of the environment, respectively.  We don't consider the level
of quantum physics here, so we assume that both agents and environments
are in one single state at every time.  While the internal state of the
agent is directly accessible to the agent, the state of the environment,
including the humans that populate it, is not directly accessible by the
agent.  The agent can only partially perceive the state of the
environment through observations.

The agent-environment pair can be represented as a pair
$(\agent,\environment)$ with $\agent=\{\agentState_t\}_{t\geq t_o}$ and
$\environment = \{\environmentState_t\}_{t\geq t_0}$ where $t_0$ is some
starting time-point.  $\agentState_t$ denotes the state of the agent at
time $t$, while $\environmentState_t$ denotes the state of the
environment at time $t$.  We emphasize once again that the environment
$\environment$ also includes the humans, and that these interact with
the agent.

\paragraph{Observations and actions.}

The agent observes the environment via sensors, and acts upon the
environment via effectors.  Every interaction between the agent and the
environment happens through these means.  Observations and actions are
considered here in a very broad sense.  Sensors provide the agent with
observations or percepts that range from low level data to natural
language input, images, movies, and so on.  Actions may range from
physical actions such as moving ahead or grasping an object, to
communicative actions (speech acts) such as displaying some data on the
screen, uttering a sentence, smiling, or playing a song or a movie.  We
do not assume that there is a synchronisation between actions and
observations, and both actions and observations can happen independently
without following a precise protocol.

To be human-centered, an HCAI agent may have to comply to restrictions
and requirements on the data that can be observed by an agent, and on
the actions that can be executed.  For instance, according to the EC
Ethics Guidelines,
%
%
observations in HCAI systems
\begin{quote}
  [\ldots] must guarantee privacy and data protection throughout a system’s
  entire lifecycle.~\cite[page 17]{eu-ethical-guidelines}
\end{quote}
and should be unbiased.  Concerning actions, the same guidelines
postulate that they should be
\begin{quote}
  [\ldots] consistent with the input, and that the decisions [to execute
    an action] are made in a way allowing validation of the underlying
  process.~\cite[page 22]{eu-ethical-guidelines}
\end{quote}
Furthermore, in the presence of an applicable legislation, an HCAI agent
should execute only legal actions, i.e., actions complying with all
applicable laws and regulations.

\paragraph{Goal directed agents.} 

AI agents always have one or more goals to achieve.%
\footnote{So-called \emph{reflex} agents \cite{aima4},
that directly map observations to actions, do not maintain an explicit
internal representation of their goals.  Most machine learning systems
belong to this category.  Even in this case, the mapping
that implements the reflexes is programmed, or learned, with reference
to some implicit objective that these agents are meant to achieve.}
An agent's goals can be codified in many different ways, depending on
the architecture of the agent.  For instance, the goal could be to
optimize a certain function, to find the best path to achieve a
position, to improve the agent's capability of recognising objects in a
scene, or to correctly answer users queries.

To be human-centered, the achievement of such a goal by an HCAI agent
should not harm the humans present in the environment.  According to the
principles of prevention of harm and fairness mentioned in the EU Ethics
Guidelines:
\begin{quote}
  AI systems should neither cause nor exacerbate harm or otherwise
  adversely affect human beings, [\dots] ensuring that individuals and
  groups are free from unfair bias, discrimination and
  stigmatisation. If unfair biases can be avoided, AI systems could even
  increase societal fairness.
  \cite[page 12]{eu-ethical-guidelines}
\end{quote}
Making explicit the goals of an HCAI agent in a necessary condition to
support these principles.
Furthermore an HCAI agent should not have goals that contrasts the goal
of the human present in the environment.

\paragraph{No full autonomy.}

HCAI agent should not be completely autonomous, and their behaviour
should always be sensitive to the external environmental or human
context.  Humans should always be allowed to intervene, e.g., to avoid
that the agent performs a harmful actions, or a specific set of them.
This is necessary to guarantee the \emph{principle of respect for human
autonomy} from the EU Ethics Guidelines~\cite[page
  12]{eu-ethical-guidelines}, that allows a human to autonomously decide
to prevent the execution of one or more actions to the HCAI agent.  From
an architectural poin of view, this means that among the set of
observations that an HCAI agent has to employ, there should be some that
detect the willingness of the humans to interrupt some planned or
executed actions.

\section{Schema for an HCAI agent}
\label{schema.sec}

\begin{figure}[h]
  \begin{center}
    \begin{tikzpicture}\sf
  \foreach \x in {0,...,4}
    \node[draw,circle,fill=yellow!40] at  (\x*360/5+54:2.9) {model$_{\x}$};
  \draw[fill=yellow!40] (0,0) circle [radius=2.75];  
  \draw[fill=red!30] (0,0) circle [radius=2];
  \draw[white,line width=10pt] (-2,0) to (2,0);
  \draw[white,line width=10pt] (0,-2) to (0,2);
  \draw (0,0) circle [radius=2];
  \foreach \x in {1,...,4}
    \coordinate (a\x) at (\x*360/4:1.5);
  \foreach \x in {0,...,4}
    \node[circle,fill=yellow!40] at  (\x*360/5+54:2.9) {model$_{\x}$};
 \node[] at  (90:2.4) {HCAI agent};
    \draw[fill=blue!70!yellow!35] (0,0) circle [radius=1.25];
    \draw[fill=green!40!black] (0,1.25) arc (90:270:1.25);
    \draw[draw=none,fill=green!40!black] (0,.6125) circle [radius=.6125];  
    \draw[draw=none,fill=blue!75!yellow!30] (0,-.6125) circle [radius=.6125];
    \node[circle,fill=green!40!black] at (0,.6125){\color{blue!75!yellow!30}\LARGE H};
    \node[green!40!black] at (0,-.5) {\LARGE E};
  \draw[draw=none,postaction={decorate,decoration={text along path,text align=center,text={Requirement}}}] (a2) to [bend left=45] (a1);
    \draw[draw=none,postaction={decorate,decoration={text along path,text align=center,text={Explanation}}}] (a3) to [bend left=45] (a2);
   \draw[draw=none,postaction={decorate,decoration={raise=-5,text along path,text align=center,text={Observation}}}] (a3) to [bend right=45] (a4);
   \draw[draw=none,postaction={decorate,decoration={text along path,text align=center,text={Action}}}] (a1) to [bend left=45] (a4);

 \end{tikzpicture}

\end{center}
\caption{A simple schematization of a Human-Centered Artificial
  Intelligence Agent (external ring).  The agent interacts with of the
  humans (H) and environment (E) (inner ring) via the elements in the
  middle ring, relying on one or several internal models.}
\label{fig:hcai-onion} 
\end{figure}
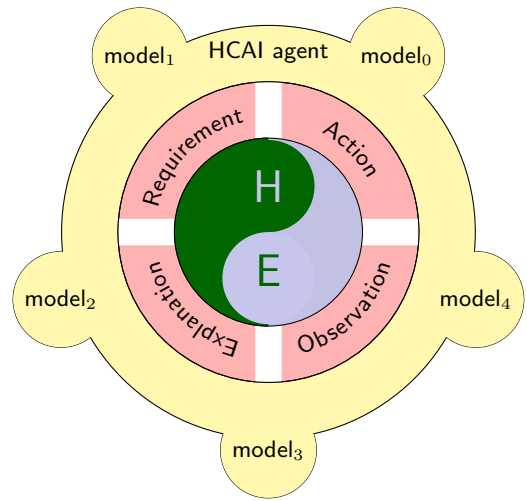

Figure~\ref{fig:hcai-onion} shows a pictorial representation of our
vision of an HCAI agent.  This picture interprets literally the
adjective ``human centered'' by drawing the human, and their
environment, at the center of the representation surrounded by the
agent.  This is the opposite of the usual representation that shows an
AI agent in the center, connected to the environment and the humans
around it.  A first observation concerns the fact that humans cannot be
separated from the environment where they operate.  Indeed they are
\emph{part} of the environment, and \emph{embedded} in it.  Humans and
environment are complementary, interconnected, and interdependent in the
natural world, and they interrelate to one another.  Therefore HCAI
Agents should take into account both humans and the environment where
humans operate.  In the picture this is represented by the blue and
green Yin-Yang symbol in the inner ring.

The outer ring in Figure~\ref{fig:hcai-onion} represents the artificial
intelligent agent.  For our discussion, the agent behaves as a unit.  In
many cases, however, the HCAI agent may be internally comprise several
interrelated intelligent agents, which interact following, e.g., a
multi-agent paradigm.  The whole HCAI agent, or any of its component
agents, accomplishes some task directly relevant for the humans it
interacts with using one or several specific approaches, or models.  For
instance, an ``artificial reasoner'' could represent its knowledge in
some logical formalism and support query answering and inference through
automatic reasoning (e.g., SAT, ASP); an ``artificial classifier'' could
be implemented in a deep neural network that is capable to classify
images into different classes; and an ``artificial planner'' can produce
plans and take decision exploiting classical planning techniques or
reinforcement learning.  To solve complex tasks, different systems and
methodologies typically need to be integrated.  A black box integration
of each ``intelligent agent'' is not sufficient; it is necessary to
integrate and make all these different approaches to collaborate one
another in a glass-box method.  We shall come back to this point when
discussing hybrid models in Section~\ref{models.classes.sec}.

The middle ring represents the interaction between the human-environment
system and the (integrated) artificial intelligent system.  The type of
interactions that one can see between human-environment system and the
artificial agent happens across a set of artifacts that are ``shared''
by the human-environment and the artificial systems.  We briefly
introduce them here, and we shall describe them more extensively in the
rest of this document.  \emph{Dependability Requirements} are artifacts,
produced by humans, that specify the expected behaviour or other
non-functional properties of the artificial intelligent system.
\emph{Actions} are considered in a broad sense.  They represent the
actions that the artificial agent can perform on the human-environment
system, as well as the action that the human can perform on the
artificial agent.  These can be physical actions, that affect the state
of the environment, or informative actions, that affect the knowledge of
the human or of the artificial agent.  \emph{Observations} are all the
data that the artificial intelligent system can collect through its
sensors.  Finally, \emph{Explanations} are artifacts produced by an
artificial agent that ``explain'' to the human the reasons of its
behaviour, e.g., the reason why it took a decision or executed an
action.  Explanations should be human understandable and acceptable in a
rational system shared by the machine and the human.


\def\observe{\mathit{Observe}}

\section{Key components of an HCAI agent} 
\label{keys.sec}

Figure~\ref{fig:hcai-onion} identifies five key components that
constitute a Human Centered AI agent: four types of artifacts that are
shared by the human-environment system and the agent and that mediate
the interactions between them; and a set of internal models.  We now
discuss these components in greater detail.

\subsection{Observations}





An agent can observe the humans and the environment through its sensors.
Here we use sensor in a broad sense, i.e., everything that can be used
by the agent to autonomously acquire data about the environment
(including humans) and everything the humans and the environment can use
to syncronously or asyncronouysy input data to the artificial agent.
Among the observations we also include the ``commands/requests'' that
humans pose to the artificial agent.

There is a great variety of types of sensors, and they all serve their
purpose depending on the information which they are meant to collect,
and on the application domain.
For example, sensors in Internet of Things applications may include
cameras, ultrasonic, infrared, acoustic, mechanical, and so on.  In
Robotics applications, sensors may include potentiometers,
accelerometers, Gimbles, lidar, an so on.  In Media applications,
sensors may include digital cameras, audio recording, textual documents,
and so on.  Essentially, sensor systems inject input data.  These can
come in streams, mostly for online systems, or in batch, mostly for
offline systems.  The way input data is represented varies widely, from
continuous to discrete, structured, lists, trees, and so on.  Input data
are typically coupled with some meta information that describes their
representation and structure.
%
In general, every item of data is associated with one or several time
stamps, that can be relative or absolute.


\def\observations{\mathbf{Obs}}
\def\observation{Obs}

\begin{definition}
  The \emph{observations} of an HCAI agent are all the data that an
  agent can aquire through its sensors.  We denote by
  $\observations_{\leq t}=\{\observation_{t'}\}_{t_0\leq t'\leq t}$ the
  observations about the human-environment system that are available to
  the agent at a given time $t$.  $\observation_t$ may or may not be a
  function of the state of the environment and the agent at time $t$.
  In general, we state that
  \[\observation_t\sim Pr(\cdot\mid\environmentState_t,\agentState_t),
  \]
  i.e., observations at time $t$ follow some (possibly unknown)
  probability distribution conditioned by the state of the environment
  and of the agent at time $t$.
\end{definition}

\subsection{Actions}

\def\actions{\mathcal{A}}

The set of actions of an agent are all the possible ways that the agent
has to affect the state of the environemnt and of the humans which it
iteracts with.  Actions can be \emph{physical actions}, e.g., robot
moves forward 10\,cm, an autonomous car brakes, an autonomous personal
assistant books a room in hotel Bellavista or buys 100 stocks of the
company SuperGulp.  Another type of action are \emph{informative
actions} that provide information to the human as a conseguence of a
decision that the agent has reached.  Examples of informative actions
are the communication of the result of a classifications of an iten
provided in input, or the communication that the price of SuperGulp
stocks is falling.

Actions should produce a tangible effect on the environment and/or the
humans in it.  This means that if execution does not fail, the state of
the humans or environment once an action is terminated should be
different from their state before the execution of that action.  Note
that an internal agent activity, like an agent taking a decision about
doing something, should not be considered as an action, because it
changes only the internal state of the agent but it does not affect the
environment.  The real action consists on the effective execution of the
decision.  Also note that in general the outcome of an action is not
deterministic.

Concrete examples of actions include the following.  If an agent's task
is to classify documents in $n$ classes, its actions are those that
\emph{communicate} the result of the classification of an item into one
or more class.  Such action will change the knowledge state of the human
that has requested the classification.  Classification without
communication should not be considered as action.  Clearly all the
actions of a robot involving physical movement are conisdered as actions
since they change the position of the robot in the environment.  The set
of actions associated to an artificial agent can be discrete or
continuous.  Notice that the set of agent's actions does not include the
actions that are executed by the humans, or other events that can
happen asynchronously in the environment.  

\def\effects{\mathbf{Eff}}
\begin{definition}
  The \emph{actions} of an HCAI agent is a set $\actions$ such that any
  $a\in\actions$ is associated to a set of effects $\effects(a)$ which
  describes a transformation of the state of the environment that is
  directly relevant for the humans in the environment.  The mapping
  $\effects(a)$ can be deterministic, non deterministic or
  probabilistic.
\end{definition}

An important aspect of the above defition is that actions in an HCAI
should be \emph{directly relevant} for the humans that populates the
environment.  This aspect relies on the pre-theoretical notion of
relevance that cannot be easily captured in a formal definition, since
relevance may emerge through complex causal chains.  In some cases,
relevance is clear: the decision of an agent to grant a mortgage to a
human is definitively relevant for the human; while the decision taken
by a system that controls the trajectory of a satellite cannot be
considered human centered, although it might have an impact on TV
viewers.
%
%
Most cases, however, are not so clear cut.  For instance, a domestic
robot's plugging to the charging station may not directly affect the
state of a human; but if this action consumes the solar-powered battery
cells of the home, this may affect the ability of the human inhabitant
to cook dinner later on.



\subsection{Explanations} 

An explanation is an explanation of some decision. In HCAI, explanations
are meant to explain to a human, or a set of humans, the decision of
executing an action $a$ made by an agent at time $t$ when it was in the
state $\agentState_t$.  As argued in~\cite{miller2019explanation}, the
definition of explanations for a decision taken by an HCAI agent cannot
be provided without making explicit reference to the human to whom the
explanation is addressed, i.e, the \emph{explanee}.  In general, an
explanation of a decision taken by the agent can be independent from how
the agent reaches such a decision.  This is especially true when the
decision is taken by a black-box method such as a neural network.

\begin{definition}[Explanation]
  An \emph{explanation} of an HCAI is a representation of the reason why
  the agent took a decision to execute a given action $a$ in state
  $\agentState_t$.  Such an explanation should be intelligible,
  understandable and acceptable by the explanee.
\end{definition}

This definition sees explanation as an artifact.  Another notion of
explanation is to see it as the process of building or communicating an
explanation.  There is a rich literature in which explanation refers to
a dialogue, and therefore a collaborative process, between the human and
the agent, where the human collaborates with the agent in the creation
of a explanation of the agent's decision that is satisfactory for the
human.
%
%
The result of this process is an explanation artifact that can be stored
and reused.

\subsection{Requirements}

A requirement, including the concept of dependability requirement, is
the formulation of a functional need that an AI system must satisfy.
This is central to the problem of verifiable AI, which has the objective
of checking that a system meets its requirements, including functional
specifications and dependability attributes.  Dependability requirements
should be expressed in a human understandable way: in fact, they are
usually formulated by humans.  The problem of verifiable AI is how to
translate these requirements in algorithms that check that the system is
compliant with them.  Being a human produced artifact, providing a sharp
definition of a requirement is rather complex if not impossible.

Some requirements may be specified in a formal language, that has an
intuitive semantics for the human, so that it is possible to verify
automatically that the artificial intelligent system behaves according
to the requirement, at least to some measurable degree of certainty.
Some requirements may not be expressed in a formal language, as
representing them in a mathematical structure is still an open issue.
Below, we concentrate on formal requirements.
  
\begin{definition}
 A formal \emph{requirement} is an expression in a formal language
 (e.g., in mathematics or logics) that unambiguously describes a
 criterion to determine if an agent-environment pair fulfils it.
\end{definition}

There are aspects of the verification of requirements for HCAI agents
that go beyond standard requirement engineering, and that stem from
these being AI agents as opposed to more traditional software systems.
In particular, AI agents have the ability to \emph{learn new knowledge}
by generalising observations done on a set of data, and to \emph{make
inferences from the learned knowledge} in order to take decision in
previously unseen or unmodeled situations.

In addition, in Human-Centered AI agents one may need to impose
requirements on \emph{how} the agent reaches the decision that leads to
a certain behaviour.  For instance, in the case of an agent that
contains machine learning models trained on data, typical requirements
concerns the fact that these data are not biased, or that they respect
the GDPR.  This type of requirements goes beyond the above definition,
that focuses on the behaviour of the agent relative to the environment
in which it operates.


\subsection{Models} 

An HCAI agent takes decisions on how to act in the environment, or how
to react to some input coming from the humans and/or the environment, on
the basis of one or more \emph{models} of the human-environment system
which it is interacting with.  Models are abstract (computational)
structures that allow to answer queries about what holds in the current
or past situations, and to predict what will be true in the future.
Models can also be used to simulate possible alternative evolutions of
the human-environment system in order to take the ``right'' decision
now.

HCAI agents are equipped with a set of models that represent the
knowledge of the agent about the human-environment system.  This
knowledge is used to support the agent in making decisions about which
actions to perform.  In general, we cannot assume that such models are
\emph{correct}, i.e., that their reflect the true state
human-environment system and that their predictions are effectively
true.  For this reason it is more appropriate to speak about
\emph{belief} instead of knowledge.  Neither we can assume that models
are \emph{complete}, i.e., that they describe the environment in all its
details.  Indeed they are \emph{simplified, abstract} representations of
some aspect of the environment obtained by abstracting away irrelevant
(or believed to be irrelevant) details.  Even at the given level of
abstraction, models may still miss information if this is not known or
observed.


\section{Models for an HCAI agent}
\label{models.sec}

The core knowledge of an HCAI agent is encapsulated in the set of models
that it adopts in order to interpret the input data, take decisions, and
provide explanations for the decisions taken.  If we ignore the physical
aspects of an agent, it is acceptable to say that the behaviour of an
HCAI agent is fully determined by its models.  Most of the requirements
and guidelines provided for HCAI agents thus concern how models are
built, how they evolve, and how they are used to take decisions.

Given this central role of models, it is useful to discuss in more
detail what are the different classes of models that we can use in an
HCAI agent, and how these models are built and how they evolve --- in
other words, the life-cycle of models.  The rest of this section is
devoted to these topics.



\subsection{Classes of AI Models}
\label{models.classes.sec}

Providing a complete and coherent classification, or even an ontology,
of AI models is beyond the scope of this paper.  For our purposes, it is
enough to list the most important, general classes of AI models.  In our
classification we take a ``technological'' perspective, i.e., we define
classes on the basis of the set of methodologies which are used to
specify the model, to represent information, and to perform decisions.
We identify four main classes plus their combination.



\paragraph{Logical Models.}  The key aspects of the environment and
  the human are represented with a logical theory (set of formulas of a
  logical/formal language) and decision on the basis of this model are
  taken via logical reasoning. Examples of this type of model are
  Logical Knowledge Bases and Ontologies, Logic programs, and planning
  domains specified via PDDL or other action languages.  Logical models
  are specified declaratively using a set of terms and formulas from a
  logic based language.  A good summary of many different logical models
  is provided in the Handbook of Knowledge
  Representation~\cite{lifschitz2008handbook}.  Logical models provide
  information of what is true, what is false and what logically follows
  from some premises.  Minker~\cite{minker2012logic} offers an overview
  of how logical models can be built and how they can be used for
  inference, decision making and planning.

\paragraph{Probabilistic Models.}  The key aspects of the environment
  and the human are represented by some probabilistic distribution.
  Decision are taken on the basis of probabilistic inference.  In an AI
  model, probability distributions are not simply defined over a set of
  hypotheses but rather over some more complex structure suitable to
  represent knowledge, as noted by Chater \textit{et
    al}~\cite{chater2006probabilistic}:
  \begin{quote}
    \it The knowledge and beliefs of cognitive agents are modeled using
    probability distributions defined over structured systems of
    representation, such as graphs, generative grammars, or predicate
    logic. This development is crucial for making probabilistic models
    relevant to cognitive science, where structured representations are
    frequently viewed as theoretically central. 
  \end{quote}
  Examples of this type of models are statistical graphical models, like
  Bayesian Networks and Hidden Markov Models.
  Usually probabilistic models distinguish between observable variables,
  which correspond to the evidence that an agent is able to observe
  directly, and hidden variables, whose distribution should be
  discovered from the data.
  The key concept in this type of model is the \emph{variable
  assignment}, i.e., an assignment to all the random variables, on which
  it is possible to apply the model in order to predict the likelihood
  of such an assignment.
  Beside probability, other mathematical theories have been used to
  build models for uncertain knowledge and plausible reasoning,
  including possibility theory~\cite{DuboisPrade.hbook2015} and belief
  function~\cite{YagerLiu.book2008}.  We extend the term ``probabilistic
  models'' to cover models based on those theories as well.

\paragraph{Real-Valued Functional Models.}  The key aspects of the worlds and
  the user are represented through (a set of) real-valued functions.
  These models can be used take in input observable quantities and
  produce an estimation of some non-observable quantity, or predictions
  of future values.  Examples of this type of models are linear models,
  support vector machines, decision trees, random forest and (deep)
  neural networks.

  A large class of real-valued functional models is constituted by
  neural network models.  A neural model is a directed graph of nodes.
  Each node is associated with a non linear activation function, the
  input of a node $n$ is a linear combination of the outputs of the
  functions associated to nodes that precede $n$ in the graph.  Both the
  linear combination and the activation function are associated with a
  set of parameters, that need to be instantiated in order to fully
  define the function.  A neural network model is also associated to a
  Loss (or Cost) function, that determines the approximation error
  (i.e., difference between known and predicted outputs) to be
  minimized.


  For every instantiation of its parameters, a neural network computes a
  function $f:R^k\rightarrow R^h$ where $k$ is the number of input nodes
  (i.e., nodes that don't have any predecessor) and $h$ is the number of
  output nodes (i.e., nodes which are not predecessor of any other
  node).  Like in all machine learning models, the main objective in a
  neural network is to find an instantiation of the parameters that
  minimizes the Loss/Cost function.

\paragraph{Dynamic Decision Models} The main purpose of these models
  is to represent the dynamics of the environment in terms of states and
  relations (transitions) between states, as well as the payoff obtained
  by the agent's being at a state.  A state represents what holds at a
  given time point in the environment; two states are connected by a
  transition if the agent can pass from one state to the other by
  executing an action.  Furthermore every state is associated with some
  ``evaluation'' function that expresses how much that state satisfies
  the objectives of the agent.  These models are used by the agent to
  decide which actions to take at every future state in order to
  maximize the likelihood of its payoff.  These models are associated
  with a set of algorithms that allow the agent to produce a ``policy''
  or a ``plan'', i.e., a sequence of actions, or some more complex
  control structure, that will reach a state with optimal (or
  sub-optimal) payoff.  This class includes a vast variety of models,
  such as (Partially Observable) Markov Decision Models, Planning
  Models, and Finite Automata.

\paragraph{Hybrid models.}  In many cases a model presents
  characteristics that are common to more than one of the classes
  described above.  We call them \emph{hybrid models}. Hybrid models are
  models that integrate some of the previous types of models.  Examples
  of such models are approaches that integrates logical and numerical
  models (e.g., Logic Tensor Networks, Lyrics) approaches that integrate
  logical and statistical models (e.g., Markov Logic Networks,
  Probabilistic Logic Programming), and approaches that integrate
  numeric, statistical and logical models (e.g., DeepProbLog, deep
  probabilistic logic programming, and Probabilistic Soft Logic).


\subsection{The lifecycle of HCAI models}

\begin{figure}
    \centering
    \begin{tikzpicture}[scale=.8,every node/.style={scale=.8},thick]
\tikzset{activity/.style={draw=red,rounded corners=10,align=center,fill=yellow!20,inner sep=5pt}}
\tikzset{data/.style={draw,align=center,fill=white,inner sep=5pt}}
  \draw[dashed,fill=green!10] (0,-2) rectangle ++(5,12);
  \draw[dashed,fill=red!10] (5,-2) rectangle ++(5,12);
  \draw[dashed,fill=blue!10] (10,-2) rectangle ++(5,12);   
  \draw (0,-2) rectangle ++(15,12);
  \node[draw,fill=white,rectangle] at (2.5,10) {Design};
  \node[draw,fill=white,rectangle] at (7.5,10) {Modelling};
  \node[draw,fill=white,rectangle] at (12.5,10) {Operating};
  \node[activity] at (5,9) (specification) {model schema \\ specification};
  \node[data] at (7.5,8) (schema) {\\ model schema \\};
  \node[builder,minimum size=1cm,left = of specification] (b) {AI engineer};
  \node[activity] at (7.5,6) (instantiation) {model instantiation \\ training, optimization, \\ knowledge encoding, \\ reinforcement learning, \\ model updating,\dots };
  \node[data] at (2.5,7) (knowledge) {supervisions \\ background knowledge \\ constraints \dots };
  \node[data] at (12.5,7) (rewards) {Rewards, feedback, \\  knowledge, \dots \\ collected at run time}; 
  \node[data] at (7.5,4) (model) {\\ Instantiated Model \\}; 
  \node[activity] at (7.5,2) (inference) {Inference with \\ the model}; 
  \node[graduate,minimum size=1cm] at (2.5,5) {knowledge expert}; 
  \node[data] at (2.5,3) (test) {Test data and \\ test cases};
  \node[data] at (12.5,3) (observations) {Observations, and \\ requests from \\ the end user};
  \node[judge,minimum size=1cm] at (2.5,1) {Requirement provider};
  \node[bob,minimum size=1cm,mirrored] at (12.5,1) {end user};
  \node[activity] at (5,0.5) (verification) {Quality control \\ end Verification};
  \node[activity] at (10,0.5) (satisfaction) {collection of \\ feedback, rewards, \\ satisfaction level, \dots};
  \node[data] at (2.5,-1) (requirements) {Requirements};
  \node[data] at (7.5,-1) (decision) {Actions, Decisions, \\ preferences \dots and \\ related explanations};
  \draw[-latex,red] (knowledge) -- (instantiation);
  \draw[-latex,red] (schema) -- (instantiation);
  \draw[-latex,red] (knowledge) -- (specification);
  \draw[-latex,red] (specification) -- (schema);
  \draw[-latex,red] (instantiation) -- (model);
  \draw[-latex,red] (model) -- (inference);  
  \draw[-latex,red] (test) -- (inference);
  \draw[-latex,red] (observations) -- (inference);  
  \draw[-latex,red] (inference) -- (decision);
  \draw[-latex,red] (decision) -- (verification);
  \draw[-latex,red] (requirements) -- (verification);
  \draw[-latex,red] (verification) -- (4,-1.5) -- (.5,-1.5) -- (.5,7) -- (knowledge);
\draw[-latex,red] (decision) -- (satisfaction);
  \draw[-latex,red] (satisfaction) -- (11,-1.5) -- (14.5,-1.5) -- (14.5,7) -- (rewards);
  \draw[-latex,red] (rewards) -- (instantiation);
\end{tikzpicture}

    \caption{AI model lifecycle}
    \label{fig:peterlifecycle}
\end{figure}
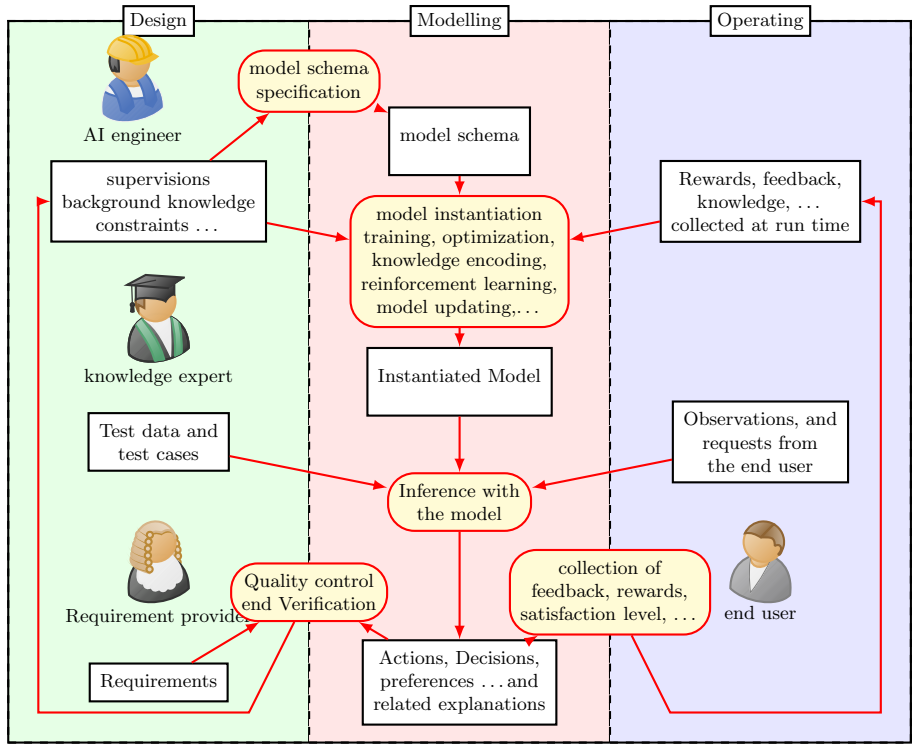

We represent the lifecycle of each model in an HCAI agent as illustrated
in Figure~\ref{fig:peterlifecycle}.  Below, we describe each main step
in this cycle.

\subsubsection{Model schema specification}

A model schema describes a class of models that share the same
structure.  The models within the same class can be obtained by
instantiating a (possibly infinite) set of parameters of the model
schema. Notice that here we use the term parameter in a broad sense,
referring both to the parameters of a probabilistic model, or a neural
network, but also to the signature of a logical language.  The model
schema specification is usually done manually, but there is a growing
interest in the community in developing methods that
(semi-)automatically learn the model structure from data: examples
include structural machine learning, programm sinthesis, non-parametric
statistical models, auto-generated neural networks, predicate invention,
and learning planning domains.

Associated to a model schema there is also an ``intuitive explanation''
of parts of the model.  For instance, in choosing a logical language to
specify the knowledge of an agent one has to say for some of the
predicates what is the intuitive meaning (with respect to states of the
environment, i.e., the proposition) that those predicates.  Similarly,
in a neural network for classifying images in N classes C1, \ldots, CN,
one has to say which of the output neuron corresponds to each class Ci.

\subsubsection{Model instance specification/learning/update}

The instantiation of a model schema amounts in setting the parameters of
the model.  This amounts in encoding a certain amount of knowledge about
the environment utilizing the ``tool-set'' provided by the model schema.
The encoding can be done manually, as it often happens in logical rule
based models via some knowledge engineering activity, or via supervised
learning from data manually labelled by humans, or in a fully
unsupervised and automatic manner (e.g., clustering).  Statistical
models can be obtained by Bayesian inference from a set of observations
or by Maximum likelihood, or maximum a posteriori inference.  Other
methods to model specifications can also be obtained by model adaptation
or transfer learning.  Similarly, updating the model can be performed
automatically via retraining, or manually by modifying the parameters.
Automatic learning of facts and rules from natural language is also
possible.  Methods for automatic learning of constraints from data are
also available.

\subsubsection{Inference with the model}

The second important aspect is how the model is used to infer a
decision, i.e., an action that the artificial agent decides to
undertake.  Given a set of observations $O$ as input, the model provides
as output a set of actions or a policy for actions.  This is obtained by
applying an \emph{inference engine} which is defined on the model.

In this phase the model instance is queried about what holds in the
environment.  Inference can be very simple, like in neural network
(simple forward propagation) or rather complex, like in constraint
satisfaction where it may be necessary to apply search or optimization
algorithms.  In logical models inference is done via some form of
logical reasoning (e.g., satisfiability) or model checking, while in
statistical models inference can be a generative process (generate a
data that has certain properties) or to compute some marginal
distribution of a certain (set of) stochastic variables.  What all the
above inference activities have in common is that they don't change the
model, but only query it.

\subsubsection{Quality Control and  Maintenance}

Once an AI artifact is ready to be used in practice, additional tasks
that are often not part of research activities become important to reach
higher Technology Readiness Levels.

For certification purposes and for the permission to use the artifact in
practice on/with non-expert users, \emph{testing and verification} procedures
for safety/security-related properties of the behavior of the model can
be necessary.
%
For products that are already in usage and needs to be updated,
\emph{maintenance and updating} methods need to exist so that problems
that are identified after shipping the model can be counteracted.
%
For supporting users and for legal reasons, it can be necessary to have
powerful methods for \emph{debugging} model properties and (re-)actions,
and for \emph{explaining} why certain outputs were (or were not)
generated.  Moreover, in connection with updates of the model, these
methods can be useful to prove to authorities that certain behavior is
excluded in the future.

This part of the AI artifact lifecycle is very relevant to human-centric
artificial intelligence, because it is the longest-lasting process in
the existence of the model where the model has contact with a large
number of untrained human individuals and unseen input data.


\section{Related work}
\label{related.sec}

With the increasing prominence of AI, there is an increasing reflection
on what it means for AI to be ``human-centered'' and several papers have
been published on this topic, some of which are discussed below.  To the
best of our knowledge, however, the work reported here is the first one
that tries to take a foundational approach to the problem of
human-centered AI, framing the discussion in technical terms by defining
the notion of a human-centered AI agent.

Wei Xu~\cite{wei2019toward} emphasizes that human centered AI systems
should exhibit behaviour which is intuitive for humans, and that this
can be achieved by adopting human centered design.  The paper proposes
an architecture that supports the collaboration between humans and
machines by considering three main factors: \emph{ethically aligned
design}, necessary to create AI agents that behave fairly, and
collaborate with humans rather than competing with them;
\emph{resemblance of human intelligence}, necessary to develop AI agents
with human-like intelligent behaviours; and \emph{comprehensibility,
usefulness and usability}, necessary to develop AI agents that are
capable of helping humans.

Ben Shneiderman~\cite{human-centered-ardito-2021} suggests that making
human-centered AI system requires the combination of AI-based
intelligent algorithms with human-centered design.  The author
highlights that, in developing HCAI, one should not limit the evaluation
to performance to the technological parts, but higher attention to human
users and other stakeholders.  This requires an increased the prominence
of user experience design and of human performance measures.  The
perspective taken in the paper is to support the 17 United Nations
Sustainable Development Goals.%
\footnote{\url{https://sdgs.un.org/goals}}
%
This work resonates with the AI model lifecycle proposed here (see
Section~\ref{models.sec}), especially in the parts of providing input
for requirement specification and of collecting the agent feedback.

In another work~\cite{shneiderman2021human}, Shneiderman stresses that
HCAI systems should give humans a greater control on the ever-increasing
automation, instead of replacing them in the decision processes.
According to the author,%
\begin{quote}
  [\ldots] humans must have \emph{meaningful control of technology} and
  are responsible for the outcomes of their actions. When humans depend
  on automation to get their work done, they must be able to anticipate
  what happens, because they, not the machines, are responsible.
\end{quote}
Shneiderman's stance is important, as it poses a limit on the decision
power of the machines and it relieves them from any responsibility.
Such a vision clashes with an opposite request of AI systems to be
increasingly autonomous.

The tension between machine autonomy and human control gave rise
recently to a field called \emph{Human Centered Machine Learning}, which
focuses on the development of HCAI systesm that adopt models that can be
trained via machine learning techniques.  See Kaluarachchi \textit{et
  al}~\cite{kaluarachchi2021review} for a review of the recent
literature.  The main objective here is to develop AI models that take
into account the input provided by humans when they get to a decision.
An example is provided by Abir De\textit{et
  al}~\cite{de2020classification}, who propose a model that classifies
images while allowing user input during inference.  An empirical
evaluation shows that this model can surpass the performance of models
trained for full automation, as well as the one of humans operating
alone.

\citeauthor{Dignum2020HowTC}~\cite{Dignum2020HowTC} argue that
human-centered AI involves a shift from an AI which is able to solve
human tasks that require some form of intelligence, to an AI which is
aware on the social environment in which it is embedded, and operates
taking into account all the limitations, the opportunities, and the
needs of the social environment.  This positions suggests that an AI
system should be considered as part of a broader \emph{socio-technical}
system.  As they put it:
\begin{quote}
  AI systems are fundamentally socio-technical, including the social
  context where it is developed, used, and acted upon, with its variety
  of stakeholders, institutions, cultures, norms and spaces.
\end{quote}
This perspective is in accordance with the schema of a human centered AI
agent proposed here (see Figure~\ref{fig:hcai-onion}), where the AI
agent is built \emph{around} the environment, and, in some sense, also
includes the social context in which it operates.


\section{Conclusions}
\label{conclusions.sec}

Human-centered AI agents are considered as part of a larger system
that also includes the humans, their society and the environment at
large. Consequently, the developments of HCAI agents is not only a
matter of developing efficient and effective altorithms to solve
complex probleems that require some form of intelligence.  HCAI looks
at the developments of AI focussing also on human values such ethical
principles, fareness, transparency of decision and objectives, \dots.
As such, human-centered AI is an intrinsically
multi-disciplinary effort that requires the establishment of a common
ground between technical and non-technical disciplines such as,
sociology, low, ethics, and philosophy.  
This paper is an attempt to offers a first step in fulfilling this requirement, by
introducing the general concept of a Human-Centered AI agent, together
with its main components and functions, as a way to bridge technical and
non-technical discussions on human-centered AI. 

\section*{Acknowledgements}

This paper has been influenced by the many discussions on the
foundations of human-centered AI held in the framework of AI4EU, the EU
landmark project to develop the European AI platform and ecosystem.  We
are indebted to the partners of work-package WP7 in AI4EU, who all
contributed to these discussions.  This work has been supported by the
European Union's Horizon 2020 research and innovation programme under
grant agreement No. 825619 (AI4EU).


\bibliographystyle{plainnat}
\bibliography{biblio}

\end{document}